\documentclass[letterpaper, 10 pt, conference]{ieeeconf}

\IEEEoverridecommandlockouts                              %

\usepackage{graphicx} %
\usepackage{amsmath}
\usepackage{amssymb}
\usepackage{booktabs}
\usepackage[table]{xcolor} %
\usepackage{multirow}
\usepackage{url}
\usepackage{hyperref}
\usepackage{xspace}

\usepackage{xfp}
\usepackage{adjustbox}
\newcommand{\APmaxchange}{13.2}

\newcommand{\apbetter}[1]{%
  \cellcolor{green!\fpeval{
    min(50,50*(#1)/\APmaxchange)
  }}+#1%
}

\newcommand{\apworse}[1]{%
  \cellcolor{red!\fpeval{
    min(50,50*(#1)/\APmaxchange)
  }}-#1%
}
\newcommand{\trainfree}{\textsuperscript{\textsc{tf}}}
\newcommand{\m}[1]{\mathcal{#1}}

\hypersetup{
    hidelinks,
    pdftitle={FunArt: Decoding Functional Structure and Articulation from Generative 3D Latents},
    }
\title{\LARGE \bf
FunArt: Decoding Functional Structure and Articulation \\
from Generative 3D Latents
}

\author{%
  Dennis Rotondi$^{1,2}$ \quad
  Abdelrhman Werby$^{1,2}$ \quad
  Kai O. Arras$^1$
 \thanks{$^1$ University of Stuttgart, Germany}
 \thanks{$^2$ International Max Planck Research School for Intelligent Systems}
}

\begin{document}

\maketitle
\thispagestyle{empty}
\pagestyle{empty}

\begin{abstract}
To operate effectively in human environments, robots must identify articulated objects, segment their movable and interactive parts, and estimate their kinematic models.
Existing articulated scene representations typically recover kinematics from observed interactions, while methods operating on static scans often decouple articulation from functional interactive elements. We present FunArt, a framework that constructs articulation-aware functional 3D scene graphs from posed RGB-D observations captured in a single static configuration. FunArt reconstructs object instances, converts their fused geometry directly into the O-Voxel representation of TRELLIS.2, and exploits its frozen, sparse-compression VAE as a structural prior. A lightweight query-based decoder combines compact object-level latents with dense, surface-aligned features to jointly segment movable parts and functional interactive elements while estimating motion type, axis, origin, and range. On the Articulate3D dataset, FunArt achieves state-of-the-art performance across movable-part segmentation, articulation estimation, and functional-element segmentation, both with and without ground-truth object input. 
In the end-to-end setting, it outperforms the strongest baselines by 1.5 $AP_{50}$ points for movable parts, 2.8 $AP_{50}$ points under joint origin-and-axis constraints, and 6.7 $AP_{50}$ points for functional elements.
These results demonstrate that generative 3D latents encode actionable structural cues that can initialize robotic perception and planning before physical interaction.
\end{abstract}

\section{Introduction}
\label{sec:introduction}

In the age of physical AI, embodied agents are moving from controlled industrial environments into homes, hospitals, and workplaces, where they must interact with articulated objects: doors swing, drawers slide, and knobs turn. Recognizing these objects and localizing them in the scene is no longer enough; agents must also identify their movable and interactive parts and infer the underlying kinematic models.
Before touching a cabinet, for example, a robot should recognize that its front is a door hinged on the left and opened by a handle on the right.
With this knowledge, it can choose an informative viewpoint, avoid the door’s swing path, and plan the pull in a single motion: capabilities unavailable if it discovers articulation only through contact.

Current approaches to acquiring this knowledge sit at two extremes.
On one side, interaction-driven methods recover articulation by \emph{watching it happen}: from classical kinematic-model estimation under manipulation~\cite{sturm2011probabilistic} to recent systems that build articulated scene representations from egocentric interaction videos or human
demonstrations~\cite{delitzas2026funrec, yu2025articulated3Dsg,
buechner2026momasg, werby2025articulated}.
These estimates are accurate precisely because motion is observed, but that is also their limit: a human must interact with every articulated object in the scene in order to capture it.
This dependence limits scalability and clashes with the goal of autonomy, in which the robot operates in a scene nobody has demonstrated.
On the other side, scene-level methods such as USDNet~\cite{halacheva2025articulate3d} predict all movable parts and their motion parameters from a static scene scan in a single shot. However, operating on the entire scene requires these methods to localize small interactive elements surrounded by large amounts of unrelated static geometry.
Instead, we first identify individual \emph{objects} and then infer their functional and movable parts, together with the corresponding motion parameters, from each object's shape.
This object-centric formulation narrows the search to relevant geometry and enables the model to learn shape-based priors about which parts serve particular functions and how they move.

Organizing inference around objects also aligns with the scene representations used by modern embodied agents: 3D scene graphs (3DSG)~\cite{armeni20193d, rotondi2026sgsurvey}  ground objects as nodes in space, and recent \emph{functional} variants attach the interactive elements objects expose~\cite{rotondi2025fungraph, zhang2025functional3dsg}.
Yet when 3DSGs are built from RGB-D images, the 3D reconstruction is used mainly to individuate object nodes and anchor them in space, while their attributes are inherited from the 2D views; the recovered geometry itself is rarely interrogated further.
Existing functional SGs consequently stop at \emph{where} functional elements are and \emph{what} they afford; how the associated parts move is either absent or, again, delegated to demonstrations.

\begin{figure}[t!]
  \centering
  \includegraphics[
  width=\linewidth,
  trim=0 0cm 0 0cm,
  clip
]{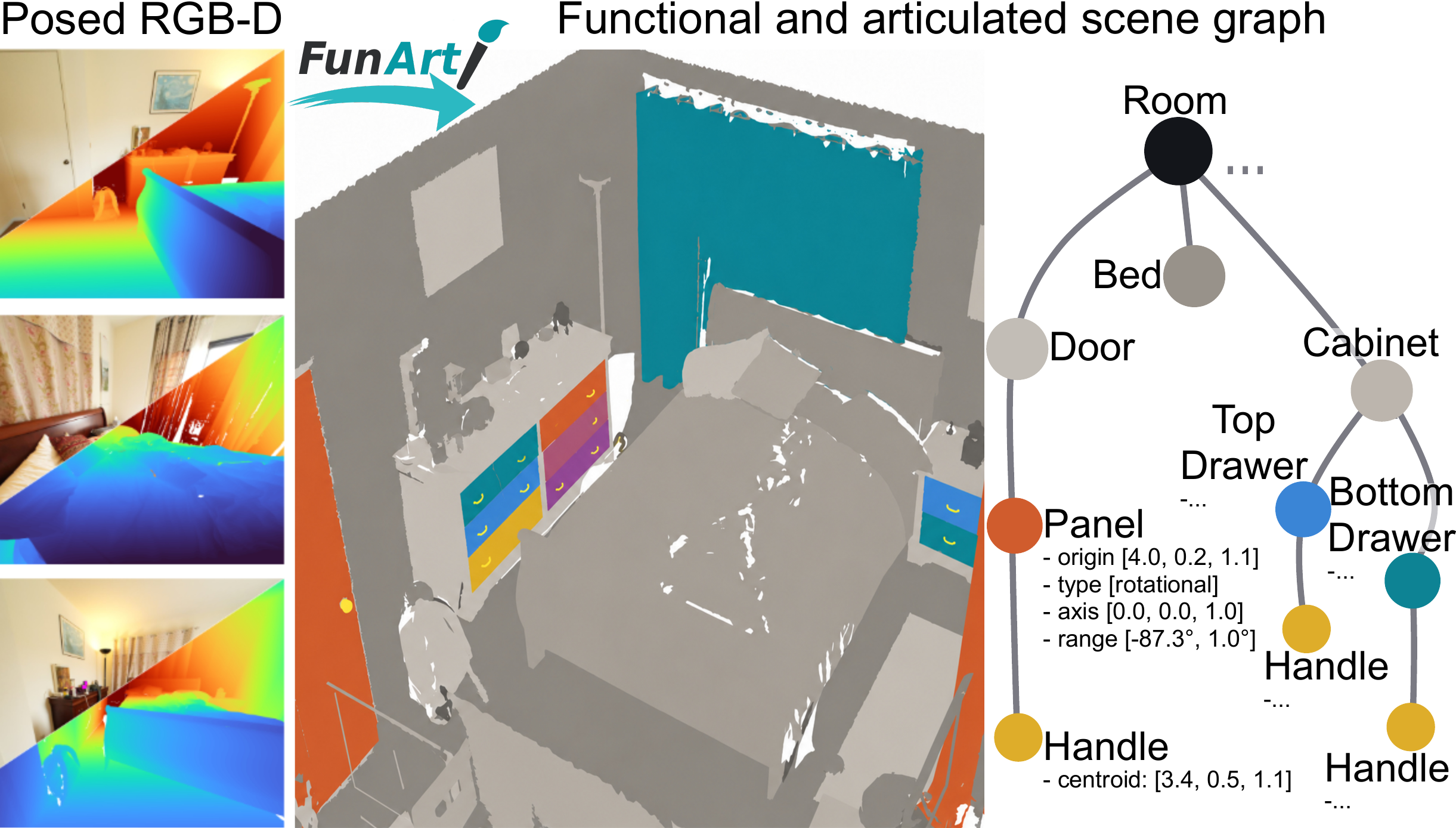}
  \caption{Given posed RGB-D observations of a scene captured in a single static configuration (left), FunArt reconstructs object instances and decomposes them into movable parts and functional interactive elements (center, colored). The predictions are assembled into a hierarchical, articulation-aware functional 3D scene graph (right): objects contain movable parts, and movable parts are associated with their interaction elements (if any). Each movable part stores its predicted motion type, axis, origin, and range, with one representative node (the door panel) shown in detail.
}
  \label{fig:splash}
\end{figure}

We close this gap by exploiting the information those graphs leave untouched, utilizing the object’s reconstructed geometry to identify movable parts and how they move.
We use the sparse 3D VAE from TRELLIS.2~\cite{xiang2025trellis2}, trained on hundreds of thousands of textured 3D assets, to learn compact latents that support high-fidelity reconstruction.
We hypothesize that this compression captures geometric and appearance regularities, such as part structure and symmetries, that also help humans infer object motion without interaction.

Concretely, we present \textbf{FunArt} (see Fig.~\ref{fig:splash}), a pipeline that turns posed RGB-D observations of a static scene into a functional 3D scene graph with articulation: an open-vocabulary instance map provides object nodes; each object is re-encoded by the frozen generative backbone, and a query-based set decoder jointly predicts nodes for movable parts and functional elements, and motion parameters that are attached to the nodes as actionable attributes.
In short, our contributions are:
\begin{itemize}
  \item We introduce the task of estimating functional scene graphs with articulation from \textit{static scene observations}, enabling autonomous inference of scene functionality and articulation without human demonstrations.
  \item We propose a modular pipeline that can be paired with any 3D instance segmenter to produce articulation-aware scene graphs capturing objects, their functional parts, and \emph{how} those parts move. At its core, a lightweight query-based decoder operates on frozen features from a pretrained large-scale 3D geometry VAE and jointly predicts movable and functional part masks together with their motion parameters.
  \item We conduct an extensive evaluation on Articulate3D, spanning movable- and interactable-part instance segmentation, articulation estimation, and settings with and without ground-truth object input.
\end{itemize}

\section{Related Work}
\label{sec:related_work}

\subsection{Articulation Estimation}

Estimating object articulation entails identifying its movable parts and characterizing their kinematics, including the joint type (revolute or prismatic), motion axis, and joint limits.
Traditionally, this problem has been addressed by detecting and tracking planar regions in dense depth images~\cite{sturm20103d} and by using a probabilistic framework to infer articulation models from object-part trajectories~\cite{sturm2011probabilistic}. 
More recent state-of-the-art methods infer articulation by leveraging segmentation and tracking foundation models~\cite{werby2025articulated} or diffusion-based denoising processes~\cite{chen2025vidbot}.

Learning-based methods can infer articulation from multiple object configurations: Ditto~\cite{ditto2022} reconstructs parts and their connecting joint from pre- and post-interaction 3D observations, while PARIS~\cite{paris23} uses multi-view images of two configurations. Recent methods extend this to articulated scene representations from recorded interactions~\cite{delitzas2026funrec, peng2025itaco}.

Although interaction and demonstration data provide strong motion cues, acquiring them requires observing the object as its articulated parts are manipulated. 
This may not be feasible in the environments in which robots are ultimately deployed.
A complementary line of work therefore estimates articulation directly from static, scene-level point clouds, building on advances in 3D deep learning~\cite{mao2022multiscan,halacheva2025articulate3d,delitzas2024scenefun3d}.
Scene-level estimation, however, must identify a relatively small number of potentially movable regions within large amounts of static geometry, which can increase computational cost and reduce accuracy in cluttered environments. 
To address this challenge, extensions~\cite{geopard25, particulate} introduce transformer-based methods to infer articulation directly from objects.
REACT3D~\cite{react3d}, meanwhile, performs zero-shot articulation estimation from a single static RGB-D observation by drawing on the general visual and semantic knowledge encoded in vision-language models (VLMs).

Nevertheless, geometry and kinematics alone may be insufficient for robotic interaction, as they do not indicate where or how a robot should interact with the object. Associating movable parts with functional interactive elements (FIEs), such as handles, knobs, and buttons, provides the actionable targets needed to execute the estimated motion.

\subsection{Functional 3D Scene Graphs}

FIEs are key to many downstream tasks~\cite{corsetti2025fun3du, corsetti2026synthfun3d, feng2026tfuns3d}, as they provide the targets through which a robot physically interacts with a scene, executing the motion predicted by an articulation model. 
3D scene graphs (3DSGs) provide a natural, explicit representation for connecting these elements to the objects they act upon. Widely used in robotics and computer vision, 3DSGs organize the geometric, semantic, and relational information of a scene to support tasks ranging from scene understanding~\cite{rana2023sayplan, rotondi2025social3dsg} to navigation~\cite{ravichandran2022hierarchical, hughes2022hydra} and manipulation~\cite{behrens2025lost}.

Following~\cite{rotondi2026sgsurvey}, a 3DSG can be represented as
\[
\m{G} \;=\; \bigl(\,\m{V},\;\; \m{V}_{\mathrm{g}},\;\; \m{V}_{\mathrm{f}},\;\; \m{E},\;\; \m{E}_{\mathrm{f}},\;\; [\,\m{L}\,]\,\bigr),
\]
where $\m{V}$ and $\m{E}$ denote the node and edge sets, $\m{V}_{\mathrm{g}}$ grounds each node in the 3D scene, and $\m{V}_{\mathrm{f}}$ and $\m{E}_{\mathrm{f}}$ assign features to nodes and edges (e.g., kinematic models), respectively. 
The function $\m{L}$ organizes the nodes into hierarchical layers~\cite{hughes2024foundations}.

Conventional 3DSGs organize scenes at multiple levels of abstraction \cite{armeni20193d, werby2024hierarchical} and primarily represent objects and their spatial or semantic relationships~\cite{wald2020learning3d, gu2024conceptgraphs}. 
Recent extensions incorporate FIEs and connect them to other entities in the scene. FunGraph~\cite{rotondi2025fungraph} detects FIEs and links them to their parent objects through intra-object relationships (edges), enabling task-driven affordance grounding. OpenFunGraph~\cite{zhang2025functional3dsg} further models both local and remote functional relationships.
KeySG~\cite{werby2025keysg} is the first framework to extend the 3DSG hierarchy from the building level down to FIEs, enriching its nodes with multimodal context extracted from RGB-D keyframes. FunFact~\cite{Fu_2026_funfact}, meanwhile, uses factor-graph reasoning to jointly infer functional relationships and quantify their uncertainty from RGB-D images.

However, these methods focus on detecting FIEs and associating them with objects without explicitly coupling them to the articulation parameters required to execute the corresponding motion. More recent approaches~\cite{buechner2026momasg,yu2025articulated3Dsg,gu2025artisg} extend 3DSGs with articulation information, but infer this information from egocentric or demonstration sequences in which articulated parts are manipulated.
To the best of our knowledge, ours is the first method to jointly estimate movable-part articulations and their associated FIEs within a functional 3DSG using only static RGB-D observations, without requiring interaction data. 
Furthermore, we show that neural refinement of the reconstructed object geometry enables accurate estimation of 3DSG node attributes.

\section{Method}
\label{sec:method}

FunArt recovers the functional and articulated structure of a static scene from a sequence of posed RGB-D observations
\begin{equation*}
\mathcal{O}
=
\{(\mathbf{I}_i,\mathbf{D}_i,\mathbf{T}_i)\}_{i=1}^{N_{\mathcal{O}}},
\end{equation*}
where $\mathbf{I}_i$, $\mathbf{D}_i$, and $\mathbf{T}_i \in SE(3)$ denote the RGB image, depth map, and camera pose of observation $i$,
respectively.
The pipeline reconstructs individual objects, converts their geometry into O-Voxel, and uses frozen TRELLIS.2 features to predict their movable parts, functional elements, and associated kinematic properties.
The output of the pipeline is a hierarchical functional scene graph $\mathcal{G}$ linking object instances to their movable parts (e.g., doors and
drawers) and functional interactive elements (e.g., handles, knobs, and buttons).
Each movable part has kinematic properties, including its motion type, origin, axis, and range.

We first introduce O-Voxel and the pretrained TRELLIS.2 geometry VAE (Sec.~\ref{sec:preliminaries}).
Next, assuming an object is given in O-Voxel form, we describe how FunArt predicts its functional and movable parts and their motion
parameters (Sec.~\ref{sec:articulation}).
Finally, we describe the direct conversion of TSDF object reconstructions into O-Voxel (Sec.~\ref{sec:method-ovoxel}).

\subsection{Preliminaries: O-Voxel Representation}
\label{sec:preliminaries}

Our method adopts the O-Voxel representation and geometric latent space introduced by TRELLIS.2~\cite{xiang2025trellis2}.
O-Voxel is a sparse voxel representation with separate shape and material features at each active cell. Its geometric component uses a flexible dual grid inspired by
dual contouring~\cite{ju2002dual}: the shape features $\mathbf{f}^{\text{shape}}_i$ of cell $\mathbf{p}_i$ encode a cell-local dual vertex $\mathbf{v}_i\in[0,1]^3$ and three edge-intersection flags $\boldsymbol{\delta}_i\in\{0,1\}^3$, which specify surface geometry and local connectivity, respectively.
The geometry VAE, pretrained at scale, compresses these features into a structured latent grid and decodes them back into O-Voxel geometry.

We use only the geometric component and its pretrained VAE, as the
articulation datasets used in this work lack ground-truth physically
based rendering (PBR) material annotations.
Hereafter, \emph{O-Voxel} and \emph{TRELLIS.2 VAE} denote this
geometric representation and its corresponding VAE.

\subsection{Articulation from Frozen Generative 3D Features}
\label{sec:articulation}

\begin{figure*}[t!]
  \centering
  \includegraphics[
  width=\linewidth,
  trim=0 2.2cm 0 2.6cm,
  clip
]{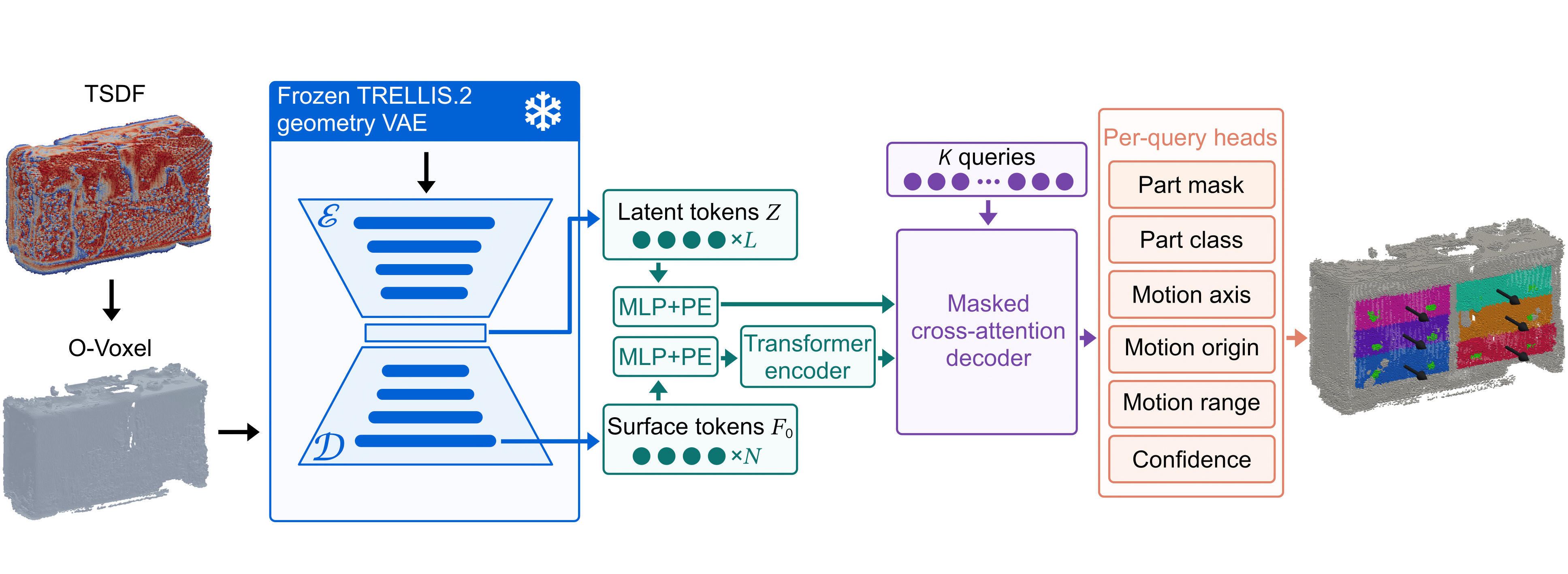}
  \caption{The input TSDF object is converted to an O-Voxel representation and processed by the frozen TRELLIS.2 geometry VAE, which provides $L$ latent tokens $Z$ and $N$ dense surface tokens $F_0$. After feature projection (via MLP) and positional encoding (PE), the surface tokens are refined by a six-layer Transformer encoder. A four-layer masked cross-attention decoder combines these features with $K$ learned queries and the latent-token context. 
  Per-query prediction heads estimate the part mask and class, motion axis, origin, and range, together with a confidence score, yielding the final proposals.}
  \label{fig:architecture}
\end{figure*}

FunArt combines features from the frozen VAE's bottleneck and decoder
to predict functional and movable parts and their kinematic properties.
Fig.~\ref{fig:architecture} provides an overview of the resulting neural network architecture.

\noindent\textit{Frozen generative geometry features.}
To make each reconstructed object compatible with TRELLIS.2, we normalize its geometry to the coordinate frame expected by the frozen backbone.
Given a point $\mathbf{x}$ in the object frame, we define
\begin{equation}
  \tau(\mathbf{x}) = s(\mathbf{x}-\mathbf{t}),
  \label{eq:canonical}
\end{equation}
where $\mathbf{t}$ is the center of the object's axis-aligned bounding box and $s$ is the inverse of its largest side length.
This transformation centers the object within the unit cube $[-\tfrac{1}{2},\tfrac{1}{2}]^3$.
We then convert the normalized geometry into a sparse O-Voxel representation at resolution $R^3=512^3$ and extract geometric features using the frozen convolutional VAE~\cite{xiang2025trellis2}.

Specifically, the VAE encoder $\mathcal{E}$ maps the O-Voxel input to a
structured latent representation (with $n_Z$ active voxels)
\begin{equation}
  Z =
  \left\{
    \left(\mathbf{z}_i,\mathbf{p}^{Z}_i\right)
  \right\}_{i=1}^{n_Z},
  \qquad
  \mathbf{z}_i\in\mathbb{R}^{32},
  \label{eq:latent_features}
\end{equation}
where $\mathbf{z}_i$ is the posterior-mean feature of active latent token $i$ and $\mathbf{p}^{Z}_i$ denotes its canonical position. 
The encoder applies a $16\times$ spatial compression, resulting in a sparse latent grid with an effective resolution of $32^3$.

The frozen decoder $\mathcal{D}$ reconstructs the O-Voxel geometry through a cascade of sparse upsampling stages at resolutions $R/16$, $R/8$, \dots, and $R$. 
The decoder re-synthesizes the surface within the input's active cells.
We extract the hidden features immediately before its final output projection:
\begin{equation}
  F_0 =
  \left\{
    \left(\mathbf{f}_i,\mathbf{p}^{F}_i\right)
  \right\}_{i=1}^{n_F},
  \qquad
  \mathbf{f}_i\in\mathbb{R}^{64},
  \label{eq:surface_features}
\end{equation}
where $\mathbf{f}_i$ is associated with an active voxel of the reconstructed surface at canonical position $\mathbf{p}^{F}_i$.

The two feature streams provide complementary geometric information. 
The bottleneck latent $Z$ captures compact object-level structure, which provides context for reasoning about part motion. In contrast, the decoder features $F_0$ retain high-resolution, surface-aligned information necessary for localizing part boundaries. 
FunArt predicts part masks from $F_0$ and uses both $F_0$ and $Z$ to classify each prediction as a rotational, prismatic, or functional part. 
For movable parts, it additionally estimates the associated kinematic parameters.

\noindent\textit{Query-based part decoding.}
Because the number of articulated and functional parts varies across objects, we formulate part discovery as direct set prediction with learned queries~\cite{carion2020detr,cheng2021mask2former}.
To form the surface attention context, we select $N{=}4096$ tokens from $F_0$ and  $L{=}256$ tokens from $Z$ using farthest-point sampling. 
Allocating tokens based on spatial coverage rather than proportional to surface area helps retain small, spatially distinct elements such as handles and switches. 
We project both feature streams to a shared dimension $d=256$ and augment them with sinusoidal positional encodings computed from their canonical positions.
The projected $F_0$ tokens are processed by a six-layer Transformer encoder~\cite{vaswani2017attention} whose self-attention relates spatially separated regions of the object surface, yielding a contextualized feature of surface voxel $i$ as $\mathbf{x}_i\in\mathbb{R}^{d}$.

We then initialize a fixed set of $K=32$ learned query embeddings,
$\{\mathbf{q}^{(0)}_k\}_{k=1}^{K}$, and process them using a four-layer Transformer decoder. The queries attend to the concatenated feature sequence $[F_0;Z]$ through masked cross-attention. Let $\mathbf{q}^{(\ell)}_k$ denote query $k$ after decoder layer $\ell$. 
At initialization and after each decoder layer, every query predicts a soft mask over the $F_0$ surface voxels. 
The mask predicted at stage $\ell$ gates the cross-attention logits over $F_0$ in layer $\ell+1$, restricting the query to the surface region assigned to it. 

The attention mask is applied asymmetrically: it gates only the surface features from $F_0$, while all latent tokens from $Z$ remain visible to every query throughout the decoder. 
The surface mask encourages each query to focus on a single part and separates nearby instances, such as multiple drawers on the same cabinet. 
At the same time, unrestricted access to $Z$ preserves the object-level context needed to infer whether and how a part can move.

We denote the final query representation by
$\mathbf{q}_k:=\mathbf{q}^{(4)}_k$. Each final query is decoded into one part
hypothesis using lightweight prediction heads~\cite{cheng2021mask2former}. 
A linear classifier predicts
\begin{equation}
  \hat{c}_k
  \in
  \left\{
    \text{rotational},
    \text{prismatic},
    \text{functional},
    \varnothing
  \right\},
\end{equation}
where $\varnothing$ denotes that the query does not correspond to a part.
Motion parameters are predicted only for rotational and prismatic parts.
We regress the motion axis directly from the final query embedding using an MLP head $g_a:\mathbb{R}^{d}\rightarrow\mathbb{R}^{3}$ and normalize it to unit length:
\begin{equation}
  \hat{\mathbf{a}}_k
  =
  \frac{g_a(\mathbf{q}_k)}
       {\lVert g_a(\mathbf{q}_k)\rVert_2}.
  \label{eq:axis_prediction}
\end{equation}
Direct regression was more stable in our preliminary experiments than differentiable SVD- or eigendecomposition-based axis extraction, particularly for thin parts with nearly degenerate geometry.
To predict the motion origin, we first estimate the center of the corresponding part. 
We convert the mask logits $\hat{m}_{k,i}$ into soft membership weights
$w_{k,i}=\sigma(\hat{m}_{k,i})$ and compute the weighted centroid
\begin{equation}
  \hat{\boldsymbol{\mu}}_k
  =
  \frac{\sum_i w_{k,i}\mathbf{p}^{F}_i}
       {\sum_i w_{k,i}}.
  \label{eq:part_centroid}
\end{equation}
An MLP $g_o:\mathbb{R}^{d}\rightarrow\mathbb{R}^{3}$ then predicts an offset from the part centroid to the motion origin:
\begin{equation}
  \hat{\mathbf{o}}_k
  =
  \hat{\boldsymbol{\mu}}_k + g_o(\mathbf{q}_k).
  \label{eq:origin}
\end{equation}
Predicting the origin relative to the part centroid is more stable than regressing its absolute position in the canonical object frame.
A separate head predicts the center $\hat{r}_k$ and half-width $\hat{h}_k$ of the motion range, giving the interval $[\hat{r}_k-\hat{h}_k,\hat{r}_k+\hat{h}_k]$. 
We also predict a confidence score that estimates the IoU of the predicted part mask, following
\cite{sun2023superpoint}.

At inference time, queries classified as $\varnothing$ are discarded, and each remaining query produces one part instance with its predicted mask and class. 
The predicted masks are propagated from the $N$
sampled surface tokens to all surface voxels by nearest-neighbor assignment in canonical coordinates.
For movable parts, the query additionally provides the corresponding kinematic parameters. The predictions are then converted from canonical coordinates to the original object frame using $\tau^{-1}$.

Finally, for each predicted component, we select the RGB frame in which its projected mask occupies the largest image area and follow FunGraph~\cite{rotondi2025fungraph} to assign an open-vocabulary label. 
To establish intra-object relationships, we create a \textit{part-of} edge from each movable component to its parent object. For each functional element, if at least $50\%$ of its mask overlaps with the mask of a movable component, we add a \textit{part-of} edge to that component.

\noindent\textit{Training objectives.}
We match the $K$ queries one-to-one with ground-truth parts using the Hungarian algorithm~\cite{kuhn1955hungarian}. 
The matching cost is a weighted sum of the negative ground-truth class probability and focal~\cite{lin2017focal} and Dice~\cite{milletari2016vnet} mask losses, using the same weights as the corresponding training terms. 
Matched queries are supervised with their assigned class and mask, whereas unmatched queries target the no-part class $\varnothing$. The classification loss $\mathcal{L}_{\mathrm{cls}}$ is a weighted cross-entropy over the three part classes and $\varnothing$, whose weight is set to $0.1$.

For a matched movable pair $(k,j)$, we use a sign-invariant loss $\ell_{\mathrm{axis}}^{k,j}$ for the axis direction and a point-to-line distance $\ell_{\mathrm{origin}}^{k,j}$for the origin.
To stabilize early training, an auxiliary linear head predicts the dominant cardinal axis ($x$, $y$, or $z$), supervised by the cross-entropy loss $\mathcal{L}_{\mathrm{card}}$ and used only during training.
Finally, $\mathcal{L}_{\mathrm{range}}$ is an unweighted $L_1$ loss on the center and half-width of the motion interval, and $\mathcal{L}_{\mathrm{score}}$ is the mean squared error between the predicted score and realized mask IoU, with unmatched queries targeting zero.

We group the losses into a part and a motion objective:
\begin{equation}
\begin{aligned}
  \mathcal{L}_{\mathrm{part}}
  ={}&\mathcal{L}_{\mathrm{cls}}
  +\lambda_f\mathcal{L}_{\mathrm{focal}}\\
  &+\lambda_d\mathcal{L}_{\mathrm{dice}}
  +\lambda_s\mathcal{L}_{\mathrm{score}},\\
  \mathcal{L}_{\mathrm{motion}}
  ={}&\lambda_a\mathcal{L}_{\mathrm{axis}}
  +\lambda_c\mathcal{L}_{\mathrm{card}}\\
  &+\lambda_o\mathcal{L}_{\mathrm{origin}}
  +\lambda_r\mathcal{L}_{\mathrm{range}},\\
  \mathcal{L}
  ={}&\mathcal{L}_{\mathrm{part}}
  +\rho\,\mathcal{L}_{\mathrm{motion}}.
\end{aligned}
\label{eq:total}
\end{equation}
We set $\lambda_f{=}5$, $\lambda_d{=}2$, $\lambda_s{=}0.5$, $\lambda_a{=}2$, $\lambda_c{=}0.5$, and $\lambda_o{=}\lambda_r{=}1$.
The motion weight $\rho$ increases linearly from $0$ to $1$ during the first ten epochs to prevent unstable early masks and assignments from disrupting motion learning.

\subsection{From Fused TSDF Reconstructions to O-Voxel Objects}
\label{sec:method-ovoxel}

Sec.~\ref{sec:articulation} assumed objects given as O-Voxel grids; we now close the loop for real scenes, where objects exist only inside a fused volumetric reconstruction. 
The standard route extracts a mesh (e.g., using marching cubes), crops the object, and rasterizes the crop using the mesh converter in~\cite{xiang2025trellis2}. 
Instead, we show that the required O-Voxel quantities can be computed directly from the interpolated TSDF, eliminating the need for an intermediate mesh.

\noindent\textit{The Instance-Labeled TSDF Map.}
In FunArt, object nodes are instantiated by an incremental instance-semantic mapping front-end inspired by OVI-Map~\cite{deng2026ovimap}. 
Class-agnostic 2D instance masks are associated across frames and fused into a global TSDF map, while open-vocabulary semantics are aggregated from a small set of automatically selected views. 
This provides temporally fused, per-instance surface geometry rather than single-frame depth back-projections.
The front-end maintains an \emph{instance-labeled TSDF map}: a sparse voxel grid with voxel size $h$, where each voxel stores a truncated signed distance $d$, an integration weight $\omega$, and an instance label.
Since the TSDF is stored only at discrete voxel centers, we extend it to a continuous field $\Phi(\mathbf{x})$ over the observed map volume by trilinearly interpolating neighboring signed-distance samples. 
We leave $\Phi(\mathbf{x})$ undefined whenever any voxel required for the interpolation is unobserved ($\omega{=}0$).
The reconstructed surface is its zero-level set:
\begin{equation}
\mathcal{S}=\{\mathbf{x}:\Phi(\mathbf{x})=0\}.    
\end{equation}

\noindent\textit{O-Voxel Geometry from the TSDF Level Set.}
TRELLIS.2's mesh converter obtains edge-intersection flags from mesh--grid intersections and places dual vertices by optimizing a quadratic error function (QEF) over intersection positions and normals. 
FunArt instead derives the edge flags from TSDF zero crossings and places dual vertices at their centroids.
$\mathbf{p}\in\{0,\dots,R-1\}^3$ indexes a grid cell, with dual vertex $\mathbf{v}(\mathbf{p})$ and edge flags $\boldsymbol{\delta}(\mathbf{p})$ as defined in
Sec.~\ref{sec:preliminaries}.

For each map instance, we extract its reconstructed surface points and remove small disconnected components. The remaining points determine the translation and scale of the canonicalization transform $\tau$ defined in Eq.~\eqref{eq:canonical}. 
We map these points to the canonical grid and restrict extraction to \emph{candidate cells}: cells containing at least one instance point, together with their two-cell neighborhood.
Corner $\mathbf{b}\in\{0,1\}^3$ of cell $\mathbf{p}$ lies at canonical position
$(\mathbf{p}+\mathbf{b})/R-\tfrac12\mathbf{1}$.
We sample $\Phi$ at its corresponding world-frame location,
\begin{equation}
  d_{\mathbf{b}}
  \;=\;
  \Phi\!\left(
    \tau^{-1}\!\left(
      \frac{\mathbf{p}+\mathbf{b}}{R}-\frac12\mathbf{1}
    \right)
  \right).
  \label{eq:corner}
\end{equation}

For each candidate cell, we evaluate $\Phi$ at its eight corners and test its twelve edges.
An edge parallel to axis $a$ connects corner $\mathbf{b}_0$ to
$\mathbf{b}_0+\mathbf{e}_a$, where $\mathbf{e}_a$ is the corresponding unit vector.
If both endpoint values $d_0$ and $d_1$ are defined and have opposite signs, we locate the zero crossing by
\begin{equation}
  \boldsymbol{\xi}
  \;=\;
  \mathbf{b}_0+\lambda\,\mathbf{e}_a,
  \qquad
  \lambda
  =
  \frac{d_0}{d_0-d_1}.
  \label{eq:crossing}
\end{equation}

A cell is active if at least one of its edges crosses the surface.
Let $Q(\mathbf{p})$ denote the set of cell-local crossings of an active cell.
We use their centroid as the dual vertex,
\begin{equation}
  \mathbf{v}(\mathbf{p})
  \;=\;
  \operatorname{clamp}_{[0,1]}
  \left(
    \frac{1}{|Q(\mathbf{p})|}
    \sum_{\boldsymbol{\xi}\in Q(\mathbf{p})}
    \boldsymbol{\xi}
  \right).
  \label{eq:dualvertex}
\end{equation}
This centroid corresponds to the mean-intersection regularization anchor in the TRELLIS.2 QEF and provides a stable dual vertex directly from the TSDF crossings.
For each canonical axis $a$, the flag $\delta_a(\mathbf{p})$ records whether the corresponding predefined edge incident to the cell's maximum-coordinate corner is crossed; the other nine edge incidences are represented by neighboring cells.

Edges touching unobserved space do not generate crossings.
Consequently, the conversion does not close unobserved regions or introduce geometry there.

\noindent\textit{Training on Reconstructed Geometry.}
The same conversion turns the reconstructions of the training scenes into a second training domain. 
We reconstruct every training scene with the mapping front-end, extract each annotated object's reconstruction (using the ground truth (GT) annotation to delimit the object, so that upstream segmentation errors do not confound the domain), encode it exactly as at inference time, and label its voxels by nearest-neighbor transfer from the GT, with a distance gate so that voxels far from any annotated surface remain background. 
These samples carry the real noise statistics of the fused geometry, including holes, over-smoothing, and lost thin structures. 

\section{Experiments}
\label{sec:experiments}
\begin{figure}[t!]
  \centering
  \includegraphics[
  width=\linewidth,
  trim=0 0cm 0 0cm,
  clip
]{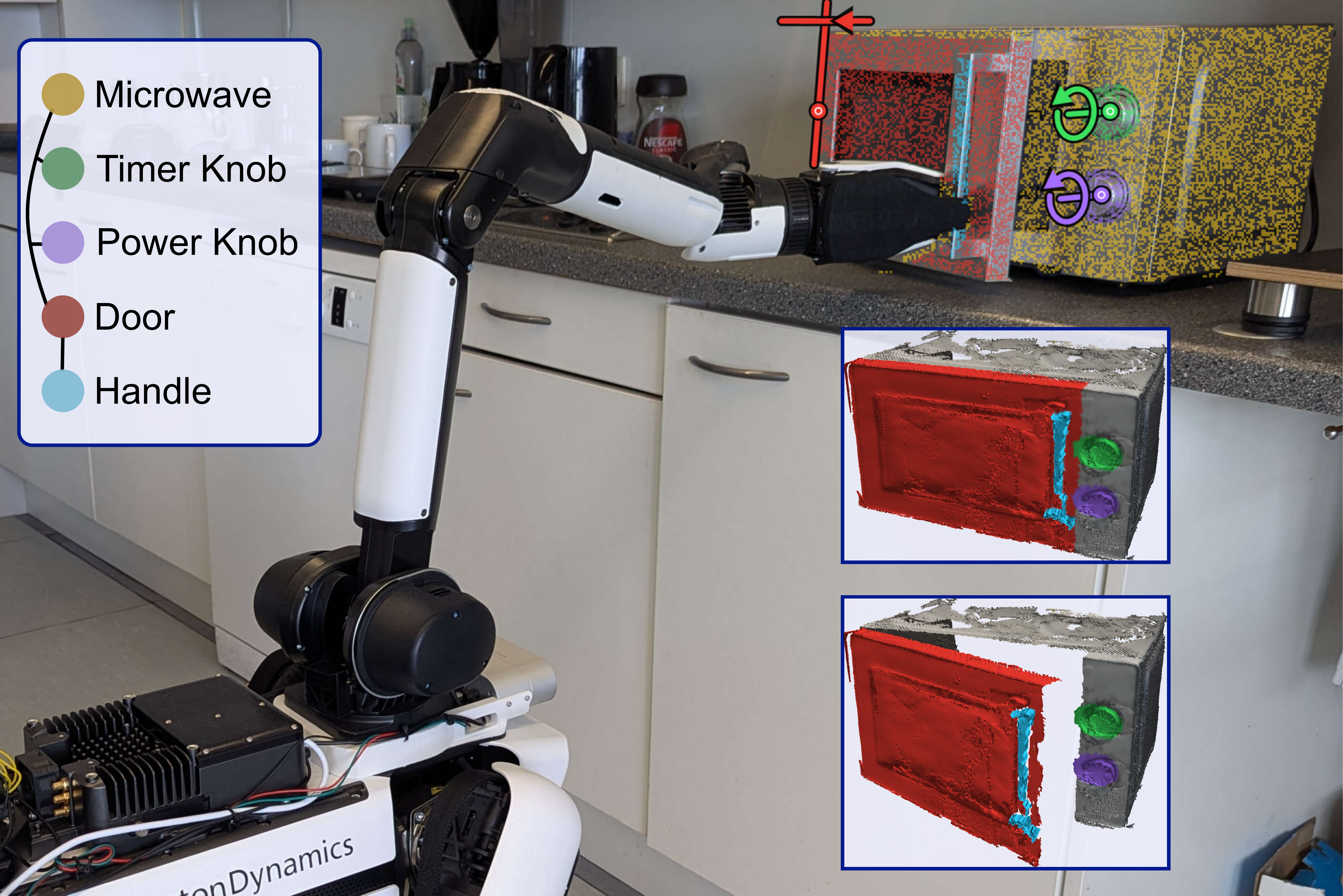}
  \caption{SPOT interacting with a real-world microwave. FunArt’s segmented 3D points and predicted motion axes are reprojected onto the RGB image. The left legend identifies the scene-graph nodes, while the insets show the 3D part segmentation in two articulation states. 
}
  \label{fig:microwave}
\end{figure}

We evaluate FunArt on movable-part instance segmentation (Sec.~\ref{sec:exp-mpis}), articulation estimation (Sec.~\ref{sec:exp-ae}), and FIE instance segmentation (Sec.~\ref{sec:exp-ieis}).
Object-to-part ownership is fixed by the object-centric formulation,
whereas FIE-to-movable-part edges are deterministically derived from
predicted mask overlap. 
Accordingly, these evaluations cover the learned quantities from which the functional and articulated scene
graph is assembled.
We further ablate our main design choices in Sec.~\ref{sec:exp-ablations}.
Fig.~\ref{fig:microwave} additionally illustrates FunArt's predicted functional and articulated structure on a real-world microwave in a robot interaction setting.

We use Articulate3D~\cite{halacheva2025articulate3d} as our benchmark. Built on ScanNet++~\cite{yeshwanthliu2023scannetpp}, it is, to the best of our knowledge, the only large-scale dataset combining 3D annotations of functional interactive elements and articulated object parts with posed RGB-D sequences of real-world scenes captured in a static state. 
All experiments use its split of 195 training and 42 validation scenes.
We report results on the full validation set under two input settings. 
The first uses GT object geometry, eliminating the effect of object detection and directly evaluating part decomposition and articulation estimation. The second uses reconstructed object geometry and therefore evaluates the complete end-to-end pipeline. 
RGB-D data is fused into TSDF volumes with a voxel size of 1 cm. 
Direct conversion to O-Voxel yields a median dual-vertex-to-surface distance of only 4.4 mm from the corresponding mesh surface, less than half the voxel size.

Open-vocabulary labels and object-instance segmentation are obtained using the existing FunGraph and OVI-MAP~\cite{deng2026ovimap} procedure and are not treated as a contribution.
All experiments are conducted on a single NVIDIA RTX~5090. The VAE is frozen, and only the 9.5M-parameter prediction head is trained. 
We use AdamW with a learning rate of $10^{-3}$, weight decay of $0.05$, a cosine learning-rate schedule, a batch size of 64, and \texttt{bfloat16} precision for 60 epochs.
Our prediction head takes an average of 10 ms per object, while converting the TSDF to the TRELLIS latent representation takes an average of 400 ms per object.

We compare against complementary classes of methods. SoftGroup~\cite{softgroup} and Mask3D~\cite{schult2023mask3d}, adapted by Articulate3D~\cite{halacheva2025articulate3d}, represent general-purpose 3D instance segmentation architectures. 
In contrast, USDNet~\cite{halacheva2025articulate3d} is a supervised method specifically designed for scene-level movable-part segmentation and articulation estimation. 
REACT3D~\cite{react3d} serves as a training-free baseline for scene-level articulation, while PARTICULATE~\cite{particulate} is an object-centric baseline that assumes GT object geometry. 
For FIE segmentation, we additionally compare with the training-free methods FunGraph~\cite{rotondi2025fungraph}, KeySG~\cite{werby2025keysg}, and FunFact~\cite{Fu_2026_funfact}, and we also adapt PARTICULATE for this task. 
Finally, FunUSD is a variant of our model that isolates the contribution of the generative features: the frozen TRELLIS.2 streams are replaced by the encoder of USDNet, a Res16UNet34C sparse-convolutional backbone~\cite{schult2023mask3d} initialized with the generic Mask3D ScanNet weights.
All supervised methods are re-trained or fine-tuned.

\subsection{Movable Part Instance Segmentation}
\label{sec:exp-mpis}
For the segmentation tasks, we adopt the official Articulate3D evaluation protocol and report average precision ($AP$), along with AP at IoU thresholds of 0.25 (${AP}_{25}$) and 0.50 (${AP}_{50}$).
As shown in Tab.~\ref{tab:movable-seg}, FunArt achieves the highest $AP$ and ${AP}_{50}$ without GT objects, running the full pipeline.
While the ${AP}_{50}$ gain over USDNet is modest, the improvement in $AP$ is more pronounced, indicating stronger performance across stricter overlap thresholds.
USDNet retains an advantage at the more permissive ${AP}_{25}$ threshold, suggesting that FunArt's primary strength lies in accurately delineating movable-part geometry.
This distinction matters for robotics: a movable-part mask specifies which observed surfaces move with a predicted joint.
When combined with estimated kinematics, accurate part geometry can help anticipate the space swept by a door or drawer, supporting collision-aware manipulation planning.

With GT objects, FunArt consistently outperforms PARTICULATE and FunUSD, confirming that the improvement also stems from stronger part decomposition and the generative features of TRELLIS.2. 
REACT3D performs poorly and exhibits a strong cabinet bias: 127 of its 144 correct part predictions belong to cabinets, although they constitute only $22\%$ of the validation objects.

\begin{table}[t]
  \centering
  \small
  \caption{Movable-part instance segmentation on Articulate3D.
  Methods are grouped according to whether ground-truth (GT) object
  segmentation is provided as input. Higher is better.
  $\dagger$ denotes a baseline adapted to this task by~\cite{halacheva2025articulate3d}. \\ \textsc{TF} denotes training-free
  methods.}
  \label{tab:movable-seg}
  \begin{tabular}{l ccc}
    \toprule
    Method & $AP \uparrow$ & $AP_{50} \uparrow$ & $AP_{25} \uparrow$ \\
    \midrule
    \multicolumn{4}{l}{\textit{Without GT object input}} \\
    SoftGroup$^\dagger$~\cite{softgroup}
      & 22.7 & 32.7 & 37.2 \\
    Mask3D$^\dagger$~\cite{schult2023mask3d}
      & 18.1 & 39.1 & 58.9 \\
    USDNet~\cite{halacheva2025articulate3d}
      & 19.8 & 41.8 & \textbf{59.9} \\
    REACT3D\trainfree~\cite{react3d}
      & 2.5 & 4.6 & 5.9 \\
    FunUSD (ours)
      & 18.1 & 33.0 & 52.2 \\
    FunArt (ours)
      & \textbf{28.5} & \textbf{43.3} & 56.8 \\
    \midrule
    \multicolumn{4}{l}{\textit{With GT object input}} \\
    PARTICULATE~\cite{particulate}
      & 16.1 & 29.2 & 46.1 \\
    FunUSD (ours)
      & 26.8 & 45.4 & 62.2 \\
    FunArt (ours)
      & \textbf{36.8} & \textbf{51.6} & \textbf{66.1} \\
    \bottomrule
  \end{tabular}
\end{table}

\begin{table}[t]
  \centering
  \small
  \caption{Articulation estimation on Articulate3D.\\  Results are
  $AP_{50}$ with the additional requirement of a correct origin,
  axis, or both. An axis is correct within $15^\circ$ and an origin
  within $0.25$\,m.}
  \label{tab:movable-motion}
  
  \begin{tabular}{l ccc}
    \toprule
    \multirow{2}{*}{Method}
      & \multicolumn{3}{c}{$AP_{50} \uparrow$} \\
    \cmidrule(lr){2-4}
      & +Origin & +Axis & +Origin+Axis \\
    \midrule
    \multicolumn{4}{l}{\textit{Without GT object input}} \\
    SoftGroup$^\dagger$~\cite{softgroup}
      & 18.5 & 21.5 & 17.7 \\
    Mask3D$^\dagger$~\cite{schult2023mask3d}
      & 24.4 & 33.8 & 19.3 \\
    USDNet~\cite{halacheva2025articulate3d}
      & 31.4 & 34.6 & 25.0 \\
    REACT3D\trainfree~\cite{react3d}
      & 3.6 & 1.8 & 1.7 \\
    FunUSD (ours)
      & 26.1 & 23.5 & 19.2 \\
    FunArt (ours)
      & \textbf{31.7} & \textbf{38.3} & \textbf{27.8} \\
    \midrule
    \multicolumn{4}{l}{\textit{With GT object input}} \\
    PARTICULATE~\cite{particulate}
      & 14.8 & 19.3 & 12.3 \\
    FunUSD (ours)
      & {34.0} & 31.9 & 30.3 \\
    FunArt (ours)
      & \textbf{38.6} & \textbf{48.7} & \textbf{37.3} \\
    \bottomrule
  \end{tabular}
\end{table}

\subsection{Articulation Estimation}
\label{sec:exp-ae}
To evaluate motion quality, we follow~\cite{halacheva2025articulate3d} and report ${AP}_{50}$ subject to additional constraints on the predicted motion origin and axis, considered both separately and jointly. An axis prediction is considered correct if its angular error is below $15^\circ$, while an origin prediction is correct if its distance from the ground truth is below 0.25~m. Tab.~\ref{tab:movable-motion} shows that FunArt performs best in both input settings. 
The gains are most pronounced for the axis-only and joint origin-and-axis criteria. 
Performance further improves when GT objects are provided, suggesting that object reconstruction remains an important bottleneck in the end-to-end setting.
\textit{Motion range.}
The Articulate3D protocol does not score motion ranges; among the compared baselines, only PARTICULATE predicts them. 
We evaluate the sign-invariant span, $\Delta r= 2*\hat{h}_k$, on GT-object inputs. A same-type prediction matched at mask IoU~$\geq0.5$ is correct within $15^\circ$/10~cm for revolute/prismatic joints. Counting unmatched parts as failures, FunArt achieves 64.9\%/37.9\% all-part recall versus 25.6\%/19.9\% for PARTICULATE. 
On matched parts, the corresponding accuracies are 87.8\%/83.7\% versus 66.8\%/47.8\%; FunArt's median errors are $3.8^\circ$/3.1~cm.

\begin{table}[t]
  \centering
  \small
  \caption{Interactable-part instance segmentation on Articulate3D.
  }
  \label{tab:interseg}
  \begin{tabular}{l ccc}
    \toprule
    Method
      & $AP \uparrow$
      & $AP_{50} \uparrow$
      & $AP_{25} \uparrow$ \\
    \midrule

    \multicolumn{4}{l}{\textit{Without GT object input}} \\
    SoftGroup$^\dagger$~\cite{softgroup}
      & 6.8 & 14.5 & 25.4 \\
    Mask3D$^\dagger$~\cite{schult2023mask3d}
      & 12.7 & 30.2 & 55.6 \\
    USDNet~\cite{halacheva2025articulate3d}
      & 12.7 & 31.1 & \textbf{55.9} \\
    \addlinespace[2pt]
    FunGraph\trainfree~\cite{rotondi2025fungraph}
      & 0.1 & 0.6 & 9.7 \\
    KeySG\trainfree~\cite{werby2025keysg}
      & 0.2 & 1.1 & 8.9 \\
    FunFact\trainfree~\cite{Fu_2026_funfact}
      & 0.1 & 0.4 & 8.5 \\
      
    FunUSD (ours)
      & 9.0 & 21.2 & 43.3 \\
    FunArt (ours)
      & \textbf{22.2} & \textbf{37.8} & 54.0 \\

    \midrule
    \multicolumn{4}{l}{\textit{With GT object input}} \\
    PARTICULATE~\cite{particulate}
      & 7.5 & 16.6 & 29.3 \\
    FunUSD (ours)
      & 14.5 & 34.7 & 61.8 \\
    FunArt (ours)
      & \textbf{26.5} & \textbf{46.4} & \textbf{67.1} \\
    \bottomrule
  \end{tabular}
\end{table}

\begin{table}[t]
  \caption{Ablations using reconstructed object extents. 
  The full model reports absolute $AP_{50}$ scores, while the remaining rows report signed $AP_{50}$-point differences relative to it.}
  \label{tab:funart-ablation}
  \centering
  \small
  \tabcolsep=0.4em

  \begin{adjustbox}{max width=0.98\columnwidth}
  \begin{tabular}{@{}lcccc@{}}
    \toprule
    \multirow{2}[2]{*}{\textbf{Variant}}
      & \multicolumn{4}{c}{\textbf{$AP_{50}\uparrow$}} \\
    \cmidrule(lr){2-5}
      & Movable
      & +Origin
      & +Axis
      & Interactive \\
    \midrule

    FunArt (full)
      & 43.3
      & 31.7
      & 38.3
      & 37.8 \\

    \midrule

    $F_0$ features only
      & \apworse{4.6}
      & \apworse{5.3}
      & \apworse{7.0}
      & \apworse{3.2} \\

    SVD-based axis estimation
      & \apworse{0.1}
      & \apbetter{0.2}
      & \apworse{9.3}
      & \apworse{0.1} \\

    Absolute origin regression
      & \apworse{0.2}
      & \apworse{7.2}
      & \apbetter{0.1}
      & \apworse{1.5} \\

    w/o $\mathcal{L}_{\mathrm{card}}$
      & \apworse{0.1}
      & 0.0
      & \apworse{4.7}
      & 0.0 \\

    w/o motion warm-up
      & \apworse{0.7}
      & \apworse{5.2}
      & \apworse{3.7}
      & \apworse{0.4} \\

    w/o reconstructed geometry
      & \apworse{12.2}
      & \apworse{10.4}
      & \apworse{13.2}
      & \apworse{11.6} \\

    \bottomrule
  \end{tabular}
  \end{adjustbox}
\end{table}

\subsection{Interactive Elements Instance Segmentation}
\label{sec:exp-ieis}
We use the same instance-segmentation metrics for FIEs. In Tab.~\ref{tab:interseg}, as for motion estimation, FunArt achieves the best ${AP}$ and ${AP}_{50}$ without GT objects and remains competitive at ${AP}_{25}$. The training-free Functional 3DSG methods struggle with this instance-level geometric evaluation because they cannot accurately reconstruct FIE from the benchmark RGB-D observations. 
In particular, our further investigation shows that FunGraph, KeySG, and FunFact detections overlap with only 49\%, 47\%, and 28\% of the GT FIEs in the validation set, respectively.
With GT objects, FunArt improves across all thresholds and clearly outperforms PARTICULATE, demonstrating its ability to accurately segment FIEs from objects, making the proposed architecture a valuable asset for refining object reconstructions for functional 3DSGs.

\subsection{Ablations}
\label{sec:exp-ablations}
Tab.~\ref{tab:funart-ablation} ablates the main design choices using the end-to-end pipeline.
Restricting the model to $F_0$ features degrades all tasks, supporting the use of the latent representation. 
SVD-based axis estimation primarily harms axis accuracy, while absolute origin regression primarily affects origin estimation, validating the proposed geometric parameterizations. Removing the dominant cardinal axis loss or motion warm-up further reduces articulation accuracy, showing their importance for stable motion learning. 
Finally, training without reconstructed geometry leads to the most severe and consistent degradation, highlighting the importance of noisy real data in the supervised signal.

\section{Conclusion}
\label{sec:conclusion}

We presented FunArt, a method for estimating articulation-aware functional 3D scene graphs from posed RGB-D observations of a scene in a single static configuration, without requiring interaction videos or demonstrations. 
Our results indicate that the frozen geometry VAE of a large 3D generative model provides structural features from which a lightweight query-based decoder can recover movable parts, functional interactive elements, and their associated kinematic parameters.
FunArt should be viewed as an (offline) initialization rather than a replacement for active perception or physical interaction. 
Predictions from static observations can remain ambiguous under occlusion or incomplete reconstruction, and the prediction head still requires annotated articulation data to generalize to new object categories and mechanisms. 
Nevertheless, FunArt provides an initial hypothesis over large scenes that can guide subsequent perception and interaction. Future work will extend the representation to richer intra- and inter-object functional dependencies and investigate how its predictions can support online active perception, collision-aware manipulation, and planning over possible scene changes.

\bibliographystyle{IEEEtran}
\bibliography{main.bib}

\end{document}